\documentclass{ztxtech}
\showlogo

\usepackage{amsmath,amsfonts,bm}

\def\Figref#1{Figure~\ref{#1}}

\def\eqref#1{equation~\ref{#1}}
\def\Eqref#1{Equation~\ref{#1}}

\def\1{\bm{1}}

\DeclareMathAlphabet{\mathsfit}{\encodingdefault}{\sfdefault}{m}{sl}
\SetMathAlphabet{\mathsfit}{bold}{\encodingdefault}{\sfdefault}{bx}{n}

\usepackage{graphicx}
\usepackage{subcaption}
\usepackage{booktabs}
\usepackage{amsmath}
\usepackage{amssymb}
\usepackage{multirow}
\usepackage{xcolor}
\usepackage{colortbl}
\usepackage{longtable}
\usepackage{hyperref}
\usepackage{url}

\newcommand{\paretoBoth}{\colorbox{blue!16}{\strut\textbf{Both}}}
\newcommand{\paretoMase}{\colorbox{orange!22}{\strut\textbf{MASE}}}
\newcommand{\bothCell}[1]{\cellcolor{blue!8}#1}
\newcommand{\maseCell}[1]{\cellcolor{orange!12}#1}

\newcommand{\method}{Fracast-0}
\newcommand{\githubrepository}{https://github.com/ztxtech/fracast-0}
\newcommand{\huggingfacerepository}{https://huggingface.co/ztxtech/fracast-0}

\def\emailicon{\raisebox{-1.5pt}{\includegraphics[height=1.05em]{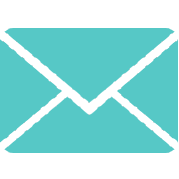}}}
\def\githubicon{\raisebox{-1.5pt}{\includegraphics[height=1.05em]{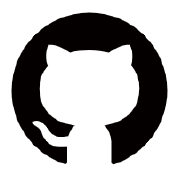}}}
\def\huggingfaceicon{\raisebox{-1.5pt}{\includegraphics[height=1.05em]{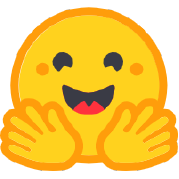}}}

\title{\method{}: Fractal Weight Sharing for a Time Series Foundation Model with Only 85K Parameters}

\hypersetup{
  pdftitle={Fracast-0: Fractal Weight Sharing for a Time Series Foundation Model with Only 85K Parameters},
  pdfauthor={Tianxiang Zhan, Huanyao Zhang, Yuanpeng He}
}

\author[1]{Tianxiang Zhan}
\author[2]{Huanyao Zhang}
\author[\ddagger,2]{Yuanpeng He}

\affiliation[1]{University of Electronic Science and Technology of China}
\affiliation[2]{Peking University}

\contribution[\ddagger]{Corresponding author}

\ztxtechdata[{\emailicon\hspace{0.3em} Email}]{\email{zhantianxianguestc@hotmail.com}}
\ztxtechdata[{\githubicon\hspace{0.3em} GitHub Repository}]{\url{\githubrepository}}
\ztxtechdata[{\huggingfaceicon\hspace{0.3em} Hugging Face Model}]{\url{\huggingfacerepository}}

\abstract{
  Time series foundation models must preserve multi-domain breadth, probabilistic output, and multiple temporal scales, but parameter count grows when each scale receives a separate representation. We introduce \method{}, a probabilistic forecasting foundation model that exploits temporal self-similarity to reuse one operator across scales. A parameter-free detector extracts significant seasonal structure. The encoder applies a shared local block along a geometric dilation ladder with scale conditioning, while the decoder combines context-gathered states with an explicit seasonal future state and reuses a second block along another ladder before emitting nine quantiles. Pretraining across six corpora preserves multi-domain breadth within 85,001 parameters. On 97 GIFT-Eval configurations without per-dataset fine-tuning, \method{} is the smallest of 28 evaluated checkpoints and remains non-dominated in the aggregate parameter-accuracy plane with MASE $0.808$ and WQL $0.564$. It uses 42.0\% fewer parameters than TinyCast, whose MASE and WQL are 4.2\% and 3.3\% lower. These results support cross-scale weight reuse as a practical route to further time series foundation model compression.}

\begin{document}
\maketitle

\section{Introduction}
\label{sec:introduction}

Time series foundation models share a simple premise. Pretraining across domains can produce forecasts without fitting each dataset. The earliest systems put that premise into different shapes, with TimesFM using a patched decoder, Moirai handling arbitrary variates and horizons with a masked encoder, and Chronos tokenizing numeric values for language-model-style pretraining \citep{timesfm,moirai,chronos}. The design space then widened as YingLong emphasized output scaling, TiRex and FlowState changed how temporal information is propagated, and Chronos-2 and Moirai-2.0 extended multivariate and decoder-only forecasting \citep{yinglong,tirex,flowstate,chronos2,moirai2}. Recent systems made these choices sharper still. Toto-2.0 examined what scaling buys, TiRex-2 carried recurrent forecasting into streaming, and Falcon-2.0 made heterogeneous pretraining regimes explicit \citep{toto2,tirex2,falcontst}. Against this increasingly specialized design space, TinyCast compressed a probabilistic zero-shot forecaster into the hundred-thousand-parameter range \citep{tinycast}. GIFT-Eval provides the common evaluation surface for these choices, covering 23 datasets and short, medium, and long horizons without per-dataset fine-tuning \citep{gifteval}.

\begin{figure}[h]
  \centering
  \includegraphics[width=\textwidth]{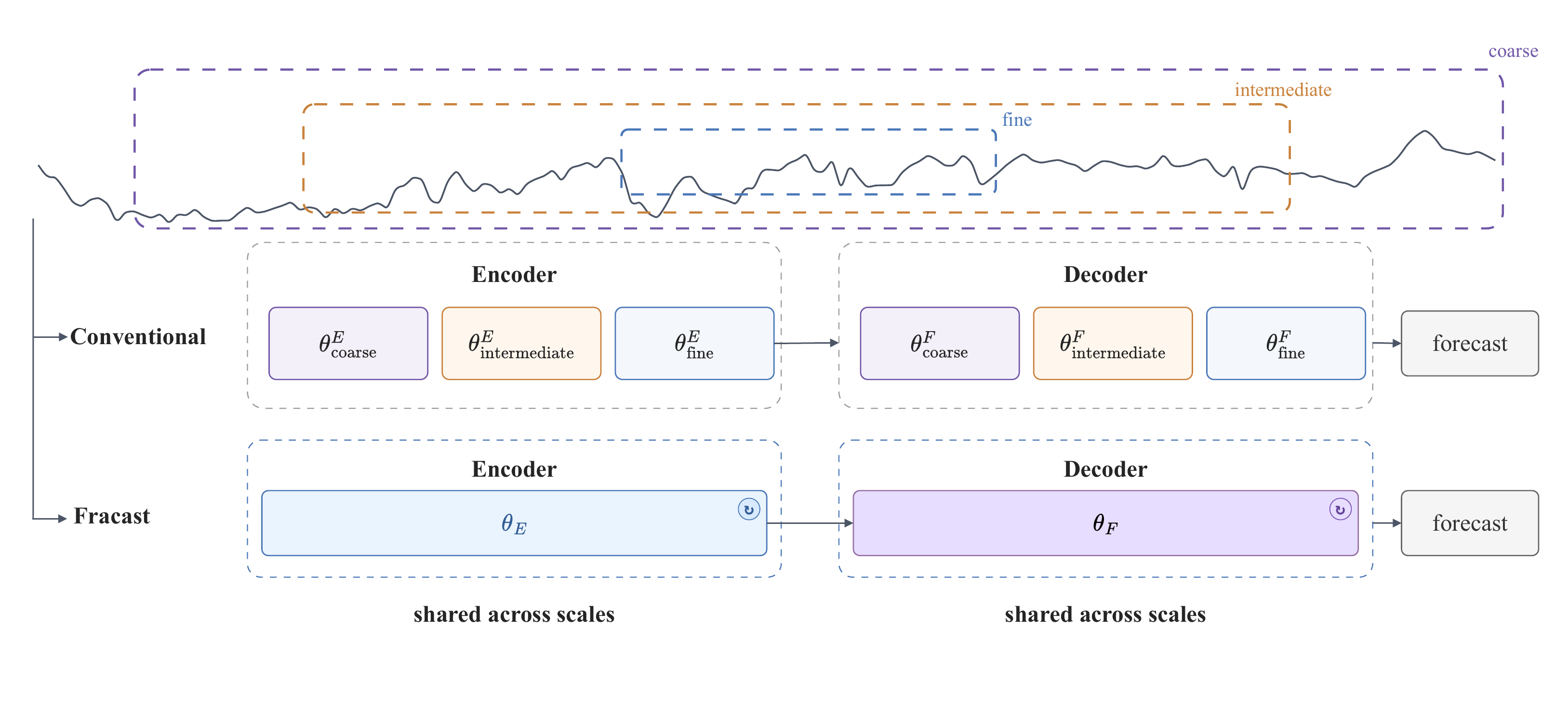}
  \caption{Cross-scale parameter growth. A conventional multiscale design assigns a separate encoder and decoder parameter set at every temporal scale. \method{} instead reuses one encoder operator and one decoder operator across scales, so the stored parameter count does not grow with the number of scale rungs.}
  \label{fig:motivation}
\end{figure}

These developments turn capacity from a background assumption into an explicit design choice. Within the compared checkpoints, TinyCast alone falls below one million parameters, using a 146K-parameter budget to compute periodic structure rather than store a large backbone. This establishes the possibility without determining a floor. Pushing below this range changes the design problem. \textbf{Breadth, probabilistic output, and multiple temporal scales must be preserved without letting parameter count grow with the number of scales.} \citep{wang2025uncertainty} \emph{Can a general probabilistic forecaster retain multi-domain evidence at a still smaller scale by reusing one operator across scales instead of assigning separate weights to each?} The two parameter layouts in \Figref{fig:motivation} make this design choice concrete. This paper pursues that question.

\method{} answers by treating temporal self-similarity as a \textbf{parameter-sharing principle}. A parameter-free detector extracts significant seasonal structure from the input. The encoder then applies one local block along a geometric dilation ladder. Scale conditioning tells the shared block which dilation it occupies, so scale changes computation rather than weights. The decoder combines context-gathered states with a seasonal future state, reuses a second local block along another dilation ladder, and emits nine quantiles. Repeated operators therefore provide multi-scale reach within an 85K-parameter budget, while pretraining across six corpora supplies breadth.

On 97 GIFT-Eval configurations without per-dataset fine-tuning, \method{} is the smallest of 28 evaluated checkpoints and remains \textbf{non-dominated} in the aggregate parameter-accuracy plane with MASE $0.808$ and WQL $0.564$ \citep{gifteval}. TinyCast uses 42.0\% more parameters and improves both aggregate metrics by 4.2\% and 3.3\%, while the \method{} gap widens from short to long horizons. Together, these results support \textbf{cross-scale weight reuse} as one practical route to a smaller general forecaster while preserving the breadth, quantile output, and multi-scale setting studied here, rather than as a universal advantage of small models.

\section{Related Works}

\paragraph{Self-similar and multiscale models.} Self-similarity lets one local rule recur at different scales. Mandelbrot framed natural length as scale-dependent. In deep learning, FractalNet uses a recursive topology, WaveNet expands receptive fields through dilation, and ALBERT ties parameters across depth \citep{mandelbrot1967,fractalnet,wavenet,albert}. Time series models adapt this principle in different ways. TimesNet aligns multi-periodic variation in 2D tensors, TimeMixer mixes coarse and fine components, and FlowState remains equivariant to sampling-rate changes \citep{timesnet,timemixer,flowstate}. Scaleformer is the closest prior method to \method{}. It iteratively refines predictions with \emph{one set of weights shared across scales} \citep{scaleformer}. \method{} shares parameters \textbf{inside the forecasting block itself}. The encoder and future-state decoder each reuse one local update along their dilation ladders, while scale conditioning separates the role of each rung.

\paragraph{Time series foundation models and GIFT-Eval.} Scale sharing is only one part of a foundation-model design. Such a model must also decide how to pretrain across domains and how to represent different variates, frequencies, and horizons. The first designs already disagreed about the representation. TimesFM used a patched decoder, Chronos tokenized numeric values for probabilistic decoding, and Moirai treated arbitrary variates and horizons with a universal transformer \citep{timesfm,chronos,moirai}. Later work extended this setup toward scaling, recurrent state, streaming, and multivariate inference \citep{toto2,tirex,tirex2,chronos2,moirai2}. GIFT-Eval includes earlier probabilistic and pretrained designs such as Lag-Llama, Timer, and TTM \citep{lagllama,timer,ttm}. Reverso and KAIROS target compact or adaptive modeling \citep{reverso,kairos}. YingLong and Falcon-2.0 explore output scaling and heterogeneous training regimes \citep{yinglong,falcontst}. TinyCast instead pushes the same setup toward a much smaller backbone \citep{tinycast}. The benchmark puts these choices on one evaluation surface with 23 datasets across domains and sampling frequencies. For each short-, medium-, or long-horizon configuration, it reports MASE and WQL relative to Seasonal Naive, and its protocol uses no per-dataset fine-tuning \citep{gifteval}. Under this protocol, accuracy is easier to interpret when read alongside model size. We therefore place \method{} at the 85K point on the parameter-accuracy frontier.

\vspace{-0.4ex}
\section{Methodology}
\label{sec:method}

\begin{figure}[h]
  \centering
  \includegraphics[width=\textwidth]{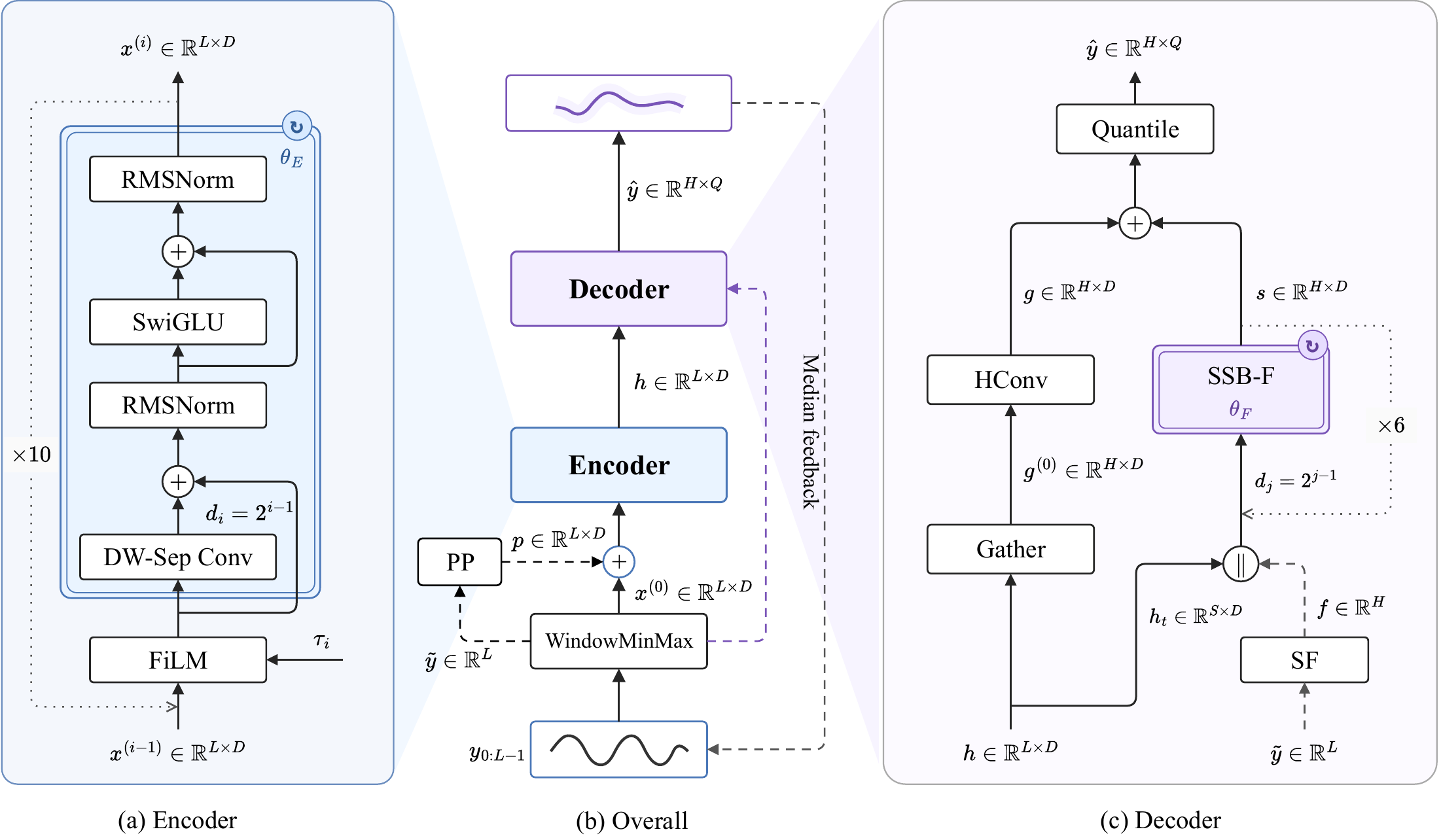}
  \caption{\method{} turns a univariate context window into a quantile forecast through three stages. A preprocessing stage normalizes the raw context and derives its periodic prior. An encoder maps the normalized context to a hidden state, and a decoder maps that state back to the horizon. Panel (b) shows the overall path. Panels (a) and (c) expand the encoder and the decoder.}
  \label{fig:architecture}
\end{figure}

\method{} processes a univariate context window in three stages, illustrated in \Figref{fig:architecture}. Channels are forecast independently with the same parameters, so a multivariate series needs no change to the architecture. The preprocessing stage normalizes the raw context and derives its periodic prior. The encoder maps the normalized context to a hidden state. The decoder maps that state to quantile forecasts over the target horizon.

\subsection{Preprocessing}

The preprocessing stage turns the raw context window into the initial state $x^{(0)}\in\mathbb{R}^{L\times D}$ that the encoder consumes. Its input is a univariate window $y\in\mathbb{R}^{L}$ together with a binary mask $m\in\{0,1\}^{L}$, where $m_t=1$ marks a position that carries a measurement and $m_t=0$ marks a gap.

\paragraph{WindowMinMax.} Series arrive at arbitrary and unrelated scales. Each window is normalized by its own range in \Eqref{eq:windowminmax}; this is \method{}'s WindowMinMax transform. RevIN \citep{revin} motivates reversible per-window normalization but uses a different mean/std rule. Let $v_{\min}=\min_t y_t$ and $v_{\max}=\max_t y_t$ be the window minimum and maximum, and let $\epsilon$ keep the denominator away from zero. The resulting $\tilde{y}\in\mathbb{R}^{L}$ lies in $[0,1]$. The two statistics are computed once per window and receive no gradient, and they are reused to invert the forecast at inference. Non-finite values are replaced by zero, and positions marked as missing in $m$ are set to zero after normalization. The same $\tilde{y}$ also feeds the seasonal fill of the decoder.

\begin{equation}
  \tilde{y}_t = \frac{y_t - v_{\min}}{\max\left(v_{\max} - v_{\min},\, \epsilon\right)},
  \label{eq:windowminmax}
\end{equation}

\paragraph{Input projection.} Each position is described by seven features. The first is the normalized value $\tilde{y}_t$. The second is a coverage channel $c_t$ that repeats the observation mask as a numeric feature, so that a missing position stays distinguishable from a measured zero. The remaining five form a bounded recency basis $r_t$, which records how far the position lies from the end of the window. Two of its channels carry the signed distance $\delta_t = (t-(L-1))/L$ linearly and logarithmically, and three are exponential decays of different rates. A single linear layer maps the concatenation $u_t = [\tilde{y}_t, c_t, r_t]\in\mathbb{R}^{7}$ to the model width $D$.

\paragraph{Periodic prior.} The seasonality of an unseen series has to be read from the context itself. \method{} uses a parameter-free detector that scores local peaks of the normalized periodogram under a Bonferroni-corrected threshold and keeps the $K$ most significant periods \citep{fisher1929}, with $K=4$ in this work; the Fisher result motivates the extremal-threshold idea rather than defining the implemented test. Every retained period $P_k$ contributes a sine and a cosine channel at each position, $\sin(2\pi t / P_k)$ and $\cos(2\pi t / P_k)$, and periods rejected by the significance test leave their channels at zero. A linear layer projects these $2K$ channels to $D$ dimensions and produces the periodic prior (PP) $p\in\mathbb{R}^{L\times D}$. The preprocessing output is the sum of the projected features and the prior, $x^{(0)} = \mathrm{Linear}(u) + p$, which is the input of the first encoder stage.

\subsection{Encoder}

The encoder maps the projected context $x^{(0)}\in\mathbb{R}^{L\times D}$ to a hidden state $h\in\mathbb{R}^{L\times D}$ that covers the whole window, and Panel (a) of \Figref{fig:architecture} shows its layout. A time series often repeats its local shape at several scales, the same statistical self-similarity that makes the measured length of a coastline depend on the ruler \citep{mandelbrot1967}. This self-similar structure is the fractal property in the name of the model, and \method{} takes it as the inductive bias of the encoder. A geometric dilation ladder gives the encoder reach over the window without any global mixing operator, and one block runs at every rung of the ladder, so the same local update applies at every scale and the parameter count does not grow with the number of scales. \method{} follows the cross-layer parameter sharing of ALBERT \citep{albert} and \textbf{ties the entire block across all ten dilations}.

\paragraph{Scale conditioning.} One block sees ten different dilations $d_i = 2^{i-1}$, so the scale has to enter through its input rather than through separate weights. \method{} conditions the block on the continuous coordinate $\tau_i = \log_2 d_i$ with the feature-wise affine transform of FiLM \citep{film} in \Eqref{eq:film},
\begin{equation}
  \begin{aligned}
     & \phi_m(\tau_i)                  = \sin(\pi \tau_i / 2^m), \quad
    \phi_{M+m}(\tau_i)              = \cos(\pi \tau_i / 2^m), \quad m = 0, \ldots, M - 1,        \\
     & [\gamma(\tau_i); \beta(\tau_i)] = W_s \phi(\tau_i) + b_s,                                 \\
     & \mathrm{FiLM}(x, \tau_i)        = x \odot \bigl(1 + \gamma(\tau_i)\bigr) + \beta(\tau_i),
  \end{aligned}
  \label{eq:film}
\end{equation}
where $M=4$ in this work, $\phi(\tau_i)\in\mathbb{R}^{2M}$ is the fixed sine/cosine scale feature \citep{vaswani}, and $W_s\in\mathbb{R}^{2D\times 2M}$ and $b_s\in\mathbb{R}^{2D}$ are initialized to zero. Conditioning therefore starts as the identity, and the shared block can fit all scales before any per-scale adjustment grows in.

\paragraph{Shared block.} Stage $i$ updates the state in \Eqref{eq:encoder-block}. One set of weights $\theta_E$ serves the whole recurrence, so the ladder is a single local operator evaluated at ten scales.
\begin{equation}
  \begin{aligned}
    a^{(i)} & = \mathrm{FiLM}\bigl(x^{(i-1)}, \tau_i\bigr), \\
    z^{(i)} & = \mathrm{RMSNorm}\Bigl(a^{(i)}
    + \mathrm{DSConv}_{d_i}\bigl(a^{(i)}\bigr)\Bigr),       \\
    x^{(i)} & = \mathrm{RMSNorm}\Bigl(z^{(i)}
    + \mathrm{SwiGLU}\bigl(z^{(i)}\bigr)\Bigr),
  \end{aligned}
  \label{eq:encoder-block}
\end{equation}
where $i=1,\dots,N$, and $\mathrm{DSConv}_{d_i}$ is the causal depthwise-separable convolution (DW-Sep Conv) \citep{mobilenets} of kernel size three whose dilation follows the ladder of WaveNet \citep{wavenet}. The depthwise stage filters every channel along time on its own, and the pointwise stage mixes channels. Both residual sums are normalized with RMSNorm \citep{rmsnorm}. The feed-forward path is a SwiGLU block \citep{swiglu}, and the whole block holds 23{,}232 parameters. Preprocessing contributes 1{,}088 parameters, while FiLM, the shared block, and the final RMSNorm contribute 24{,}448 to the encoder, so the full core holds 25{,}536 parameters. This final RMSNorm turns $x^{(N)}$ into the hidden state $h$.

\subsection{Decoder}

The decoder maps $h\in\mathbb{R}^{L\times D}$ to $\hat{y}\in\mathbb{R}^{H\times Q}$ for $H=48$ steps and $Q=9$ quantile levels; Panel (c) of \Figref{fig:architecture} shows its layout. The gather state $g$ summarizes the encoded context at every future position, while the future-state residual $s$ evolves causally from a seasonal reference. Their sum determines the quantiles.

\paragraph{Gather path.} This path reduces the encoded context to a shared summary, gives each horizon position an explicit query, and refines the positions causally together. Let $\bar{h}=L^{-1}\sum_{\ell=1}^{L}h_\ell$ and $c=[\bar{h};h_L]\in\mathbb{R}^{2D}$. For $j=L,\ldots,L+H-1$, the joint projection and horizon refinement are
\begin{equation}
  z_j = W_g[r_j;c]+b_g+u_j,\qquad
  g_j=\mathcal{H}_2\bigl(\mathcal{H}_1(\mathcal{S}(z_j))\bigr),
  \label{eq:gather}
\end{equation}
where $r_j\in\mathbb{R}^{5}$ is the bounded recency feature, $u_j\in\mathbb{R}^{D}$ is the learned position query, and $W_g\in\mathbb{R}^{D\times(2D+5)}$ and $b_g$ form a linear layer. The normalized feed-forward update and horizon refinement map are
\begin{equation}
  \mathcal{S}(z)=\mathrm{RMSNorm}\bigl(z+\mathrm{SwiGLU}(z)\bigr),
  \label{eq:gather-ffn}
\end{equation}
\begin{equation}
  \mathcal{H}_\ell(z)=\mathrm{RMSNorm}\bigl(z+\mathrm{Conv}_{2^{\ell-1}}(z)\bigr),
  \quad \ell=1,2,
  \label{eq:gather-hconv}
\end{equation}
where each convolution is causal, depthwise, and of kernel size three. The output is $g\in\mathbb{R}^{H\times D}$.

\paragraph{Future-state path.} The gather path carries context only through a pooled summary, so this path injects an explicit seasonal state and evolves it causally over the horizon. The detector chooses the primary period $P_1$ among up to $K=4$ significant normalized-periodogram candidates, or $P_1=1$ when none is significant. Each future position $t$ enters one of $B=16$ phase bins $b(t)=\lfloor B(t\bmod P_1)/P_1\rfloor$; the seasonal fill in that bin is the mean of observed normalized context values. An empty bin uses the context mean, so the fill $f_j$ is deterministic. It is projected with recency, $p_j=W_f[f_j;r_j]+b_f$, and the shared state begins as $z_F^{(0)}=[h_{L-S+1:L};p_1;\ldots;p_H]$, where $S=128$. Let $\mathcal{F}_6$ denote six applications of the decoder-specific SSB-F block with dilations $1,2,4,8,16,32$. Each application composes a causal kernel-three depthwise-separable convolution, residual RMSNorm, SwiGLU, and RMSNorm, without FiLM. The residual state is
\begin{equation}
  s = W_F\bigl[\mathcal{F}_6\bigl(z_F^{(0)}\bigr)\bigr]_{S+1:S+H}+b_F,
  \qquad W_F=0,\; b_F=0,
  \label{eq:future}
\end{equation}
where only the final $H$ states are read and $s\in\mathbb{R}^{H\times D}$; zero initialization makes the path an identity residual at the start of training.

\paragraph{Combination and rollout.} The decoder computes $\hat{y}=W_o(g+s)+b_o$, where $W_o\in\mathbb{R}^{Q\times D}$ and $b_o\in\mathbb{R}^{Q}$. WindowMinMax statistics invert the normalized forecast at inference. Longer horizons feed the median forecast back into the context and rerun the same model without new parameters. The gather path with output projection holds 31{,}625 parameters, the future-state path holds 27{,}840, the decoder holds 59{,}465, and the model holds 85{,}001 in total.

\subsection{Training}

\method{} is trained on univariate windows in the normalized space. For normalized targets $y_t$, predicted quantiles $\hat{q}_t(\tau)$, observation mask $o_t$, and $\mathcal{Q}=\{0.1,0.2,\ldots,0.9\}$, the pinball loss of quantile regression \citep{koenker1978regression,timesfm} is
\begin{equation}
  \mathcal{L}_{\mathrm{pin}}
  = \frac{1}{|\mathcal{Q}|}\sum_{\tau\in\mathcal{Q}}
  \frac{1}{\sum_{t=1}^{H}o_t}\sum_{t=1}^{H} o_t
  \max\left((\tau-1)(y_t-\hat{q}_t(\tau)),\,
  \tau(y_t-\hat{q}_t(\tau))\right),
  \label{eq:pinball}
\end{equation}
where positions with $o_t=0$ and near-flat windows are excluded. The committing term uses the median error $m_t=|\hat{q}_t(0.5)-y_t|$ and the error $c_t=|s_t-y_t|$ of the seasonal-copy reference $s_t$. Let $\gamma=1$ when $\sum_{t=1}^{H}o_t c_t < \sum_{t=1}^{H}o_t m_t$, and $\gamma=0$ otherwise:
\begin{equation}
  \mathcal{L}_{\mathrm{commit}}
  = \gamma\,
  \frac{\sum_{t=1}^{H} o_t \max(0,m_t-c_t)}{\sum_{t=1}^{H} o_t},
  \label{eq:commit}
\end{equation}
so each positive term penalizes a position where the median is less accurate than the seasonal copy, and $\gamma$ disables the whole term when the copy is not better over the window. The total objective is
\begin{equation}
  \mathcal{L} = \mathcal{L}_{\mathrm{pin}} + \lambda \mathcal{L}_{\mathrm{commit}},
  \label{eq:total-loss}
\end{equation}
with $\lambda=0.3$ reported in Appendix~\ref{app:detail}. Training unrolls several future chunks and feeds each chunk with either the model median or the observed continuation under scheduled sampling that increases from zero, matching the objective to long-horizon rollout. Exact rollout length, sampling schedule, optimizer settings, and budget are reported in Appendix~\ref{app:detail}.

\raggedbottom
\section{Experiments}
\label{sec:experiments}

\paragraph{Benchmark.} We evaluate forecasting on GIFT-Eval \citep{gifteval}, a benchmark of 23 datasets across seven domains and ten sampling frequencies with short, medium, and long horizon terms, without per-dataset fine-tuning. We follow its official test protocol with the released windows and the nine quantile levels from $0.1$ to $0.9$. Point accuracy is the mean absolute scaled error (MASE) normalized by the Seasonal Naive score of the same configuration, and probabilistic accuracy is the weighted quantile loss (WQL); both are aggregated with a geometric mean over the $97$ configurations, of which $55$ are short, $21$ medium, and $21$ long. The scaling study compares released checkpoints of Moirai and Moirai~2.0 \citep{moirai,moirai2}, TimesFM \citep{timesfm}, Chronos-Bolt and Chronos-2 \citep{chronos,chronos2}, Toto~2.0 \citep{toto2}, TiRex and TiRex-2 \citep{tirex,tirex2}, FlowState \citep{flowstate}, KAIROS \citep{kairos}, \mbox{YingLong} \citep{yinglong}, Falcon-2.0 \citep{falcontst}, and TinyCast \citep{tinycast}; their scores are leaderboard entries recomputed under the same aggregation.

\paragraph{Pretraining data.} \method{} is pretrained on six public corpora: the GIFT-Eval pretraining corpus \citep{gifteval}, LOTSA, the Chronos datasets, BOOM, FEV, and the synthetic shards released with TinyCast \citep{tinycast}. No GIFT-Eval train split enters the corpus, and windows are drawn from all sources with a band-mix policy over sampling-rate bands.

\paragraph{Parameter budget.} \method{} reduces its budget by exploiting the self-similarity of temporal structure along the dilation ladder: one local block is reused at every scale, so the parameter count does not grow with the number of rungs. The largest single reduction is the periodic prior ($-96.5\%$), while the main sharing gain is the encoder's dilated convolutions ($-59.9\%$); the future-state path saves another $37.9\%$. These reductions outweigh the extra parameters in FiLM scale conditioning and the wider gather head. TinyCast \citep{tinycast} provides a compact reference at this scale and shares the parameter-free detector, bounded recency basis, and seasonal fill. Table~\ref{tab:budget} compares the two budgets at the same context length $L=2048$, block horizon $H=48$, quantile count $Q=9$ and width $D=64$, and records where each model spends its budget rather than only what it saves.

\begin{table}[h]
  \centering
  \caption{TinyCast and \method{} parameter budgets, counting shared tensors once.}
  \label{tab:budget}
  \resizebox{\textwidth}{!}{%
    \renewcommand{\arraystretch}{1.05}%
    \setlength{\tabcolsep}{25pt}%
    \setlength{\fboxsep}{1pt}%
    \begin{tabular}{@{}lrrrr@{}}
      \toprule
      Component                     & TinyCast           & \method{} (Ours)  & Difference         & Relative (\%)                     \\
      \midrule
      Input projection              & 960                & 512               & -448               & \colorbox{red!14}{-46.7}          \\
      Periodic prior                & 16{,}448           & 576               & -15{,}872          & \colorbox{red!24}{-96.5}          \\
      Scale conditioning            & --                 & 1{,}152           & +1{,}152           & --                                \\
      \addlinespace[1.2ex]
      Encoder, dilated convolutions & 57{,}920           & 23{,}232          & -34{,}688          & \colorbox{red!20}{-59.9}          \\
      Encoder, final normalization  & --                 & 64                & +64                & --                                \\
      \addlinespace[1.2ex]
      Decoder, query and gather     & 26{,}313           & 31{,}625          & +5{,}312           & \colorbox{green!18}{+20.2}        \\
      Decoder, future-state path    & 44{,}864           & 27{,}840          & -17{,}024          & \colorbox{red!12}{-37.9}          \\
      \midrule
      \textbf{Total}                & \textbf{146{,}505} & \textbf{85{,}001} & \textbf{-61{,}504} & \colorbox{red!18}{\textbf{-42.0}} \\
      \bottomrule
    \end{tabular}%
  }
\end{table}

\begin{figure}[h]
  \centering
  \begin{minipage}[t]{0.49\textwidth}
    \centering
    \includegraphics[width=\linewidth]{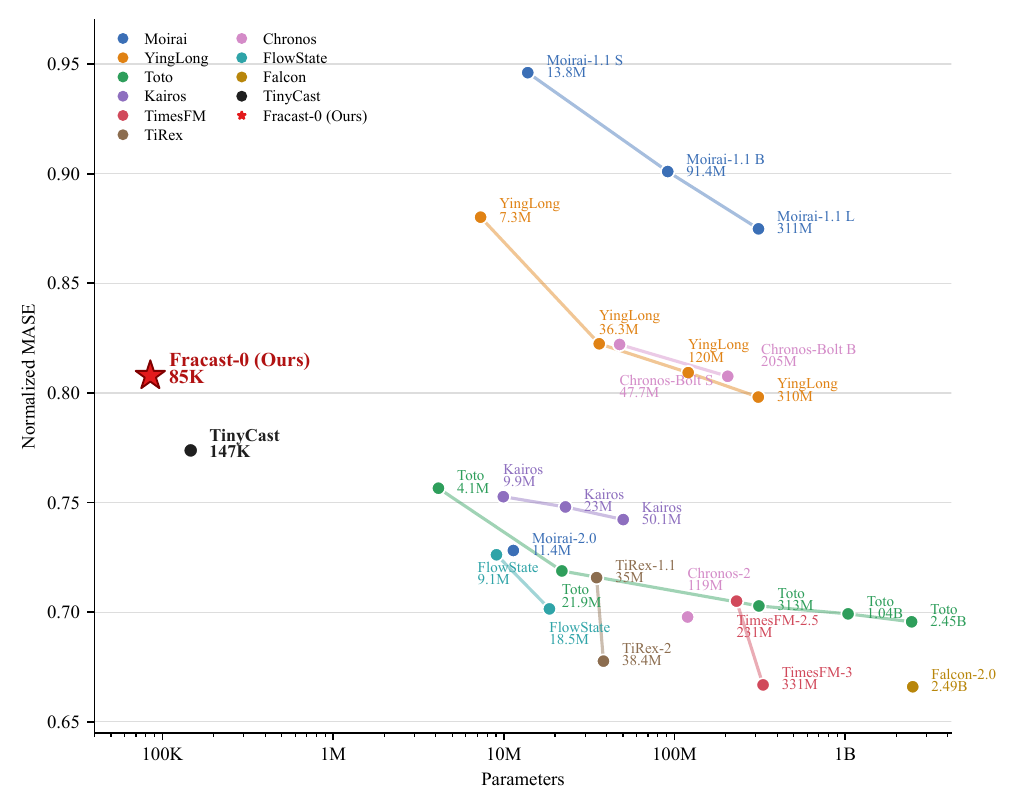}
    \centerline{\small (a) Normalized MASE}
  \end{minipage}\hfill
  \begin{minipage}[t]{0.49\textwidth}
    \centering
    \includegraphics[width=\linewidth]{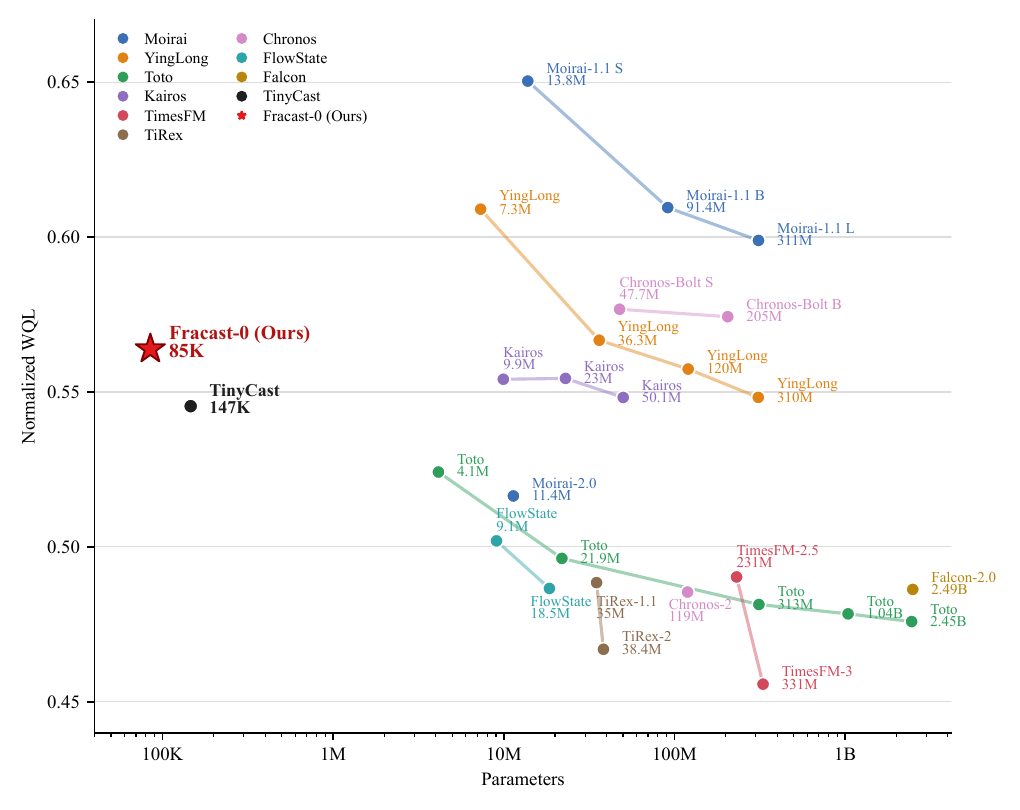}
    \centerline{\small (b) Normalized WQL}
  \end{minipage}
  \caption{Scaling study on GIFT-Eval. Every point is a released checkpoint, and lines connect checkpoints of the same family.}
  \label{fig:scaling}
\end{figure}

\begin{table}[h]
  \centering
  \caption{GIFT-Eval test results without per-dataset fine-tuning. MASE and WQL are Seasonal
    Naive-normalized geometric means over 97 configurations (55 short, 21 medium, 21 long).}
  \label{tab:main}
  \resizebox{\textwidth}{!}{%
    \renewcommand{\arraystretch}{1.05}%
    \setlength{\tabcolsep}{12pt}%
    \begin{tabular}{@{}lrr*{8}{r}@{}}
      \toprule
                                &        &             & \multicolumn{4}{c}{Normalized MASE} & \multicolumn{4}{c}{Normalized WQL}                                                                                                                   \\
      \cmidrule(lr){4-7}\cmidrule(lr){8-11}
      Model                     & Params & Pareto      & Short                               & Medium                             & Long             & Overall
                                & Short  & Medium      & Long                                & Overall                                                                                                                                              \\
      \midrule
      \textbf{\method{} (Ours)} & 85K    & \paretoBoth & \bothCell{0.770}                    & \bothCell{0.844}                   & \bothCell{0.877} & \bothCell{0.808} & \bothCell{0.569} & \bothCell{0.557} & \bothCell{0.556} & \bothCell{0.564} \\
      TinyCast                  & 147K   & \paretoBoth & \bothCell{0.755}                    & \bothCell{0.786}                   & \bothCell{0.811} & \bothCell{0.774} & \bothCell{0.570} & \bothCell{0.515} & \bothCell{0.515} & \bothCell{0.545} \\
      Moirai-2.0                & 11.4M  & --          & 0.695                               & 0.759                              & 0.789            & 0.728            & 0.517            & 0.519            & 0.513            & 0.516            \\
      FlowState-r1.1            & 18.5M  & \paretoBoth & \bothCell{0.682}                    & \bothCell{0.720}                   & \bothCell{0.736} & \bothCell{0.701} & \bothCell{0.507} & \bothCell{0.469} & \bothCell{0.455} & \bothCell{0.487} \\
      TiRex-2                   & 38.4M  & \paretoBoth & \bothCell{0.662}                    & \bothCell{0.692}                   & \bothCell{0.706} & \bothCell{0.678} & \bothCell{0.489} & \bothCell{0.443} & \bothCell{0.436} & \bothCell{0.467} \\
      Chronos-2                 & 119M   & --          & 0.667                               & 0.725                              & 0.757            & 0.698            & 0.496            & 0.471            & 0.472            & 0.485            \\
      YingLong-310M             & 310M   & --          & 0.758                               & 0.841                              & 0.867            & 0.798            & 0.562            & 0.535            & 0.526            & 0.548            \\
      TimesFM-3                 & 331M   & \paretoBoth & \bothCell{0.648}                    & \bothCell{0.685}                   & \bothCell{0.700} & \bothCell{0.667} & \bothCell{0.475} & \bothCell{0.432} & \bothCell{0.430} & \bothCell{0.456} \\
      Toto-2.5B                 & 2.45B  & --          & 0.663                               & 0.726                              & 0.754            & 0.696            & 0.486            & 0.464            & 0.461            & 0.476            \\
      Falcon-2.0                & 2.49B  & \paretoMase & \maseCell{0.652}                    & \maseCell{0.667}                   & \maseCell{0.703} & \maseCell{0.666} & 0.492            & 0.474            & 0.483            & 0.486            \\
      \bottomrule
    \end{tabular}%
  }
\end{table}

\begin{figure}[h]
  \centering
  \includegraphics[width=\textwidth]{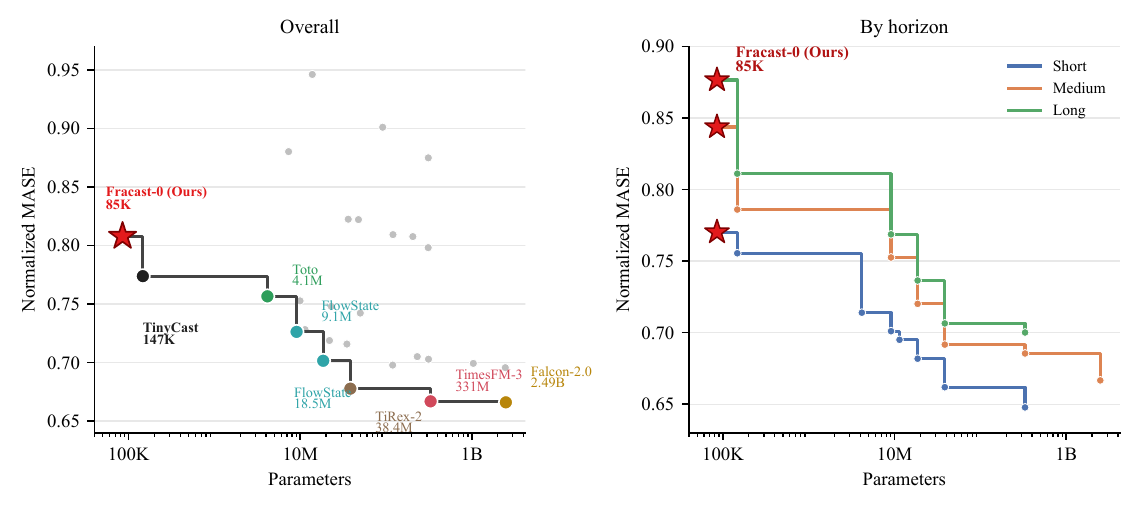}
  \caption{Parameter-accuracy Pareto frontier on GIFT-Eval.}
  \label{fig:pareto}
\end{figure}

\begin{figure}[h]
  \centering
  \includegraphics[width=\textwidth]{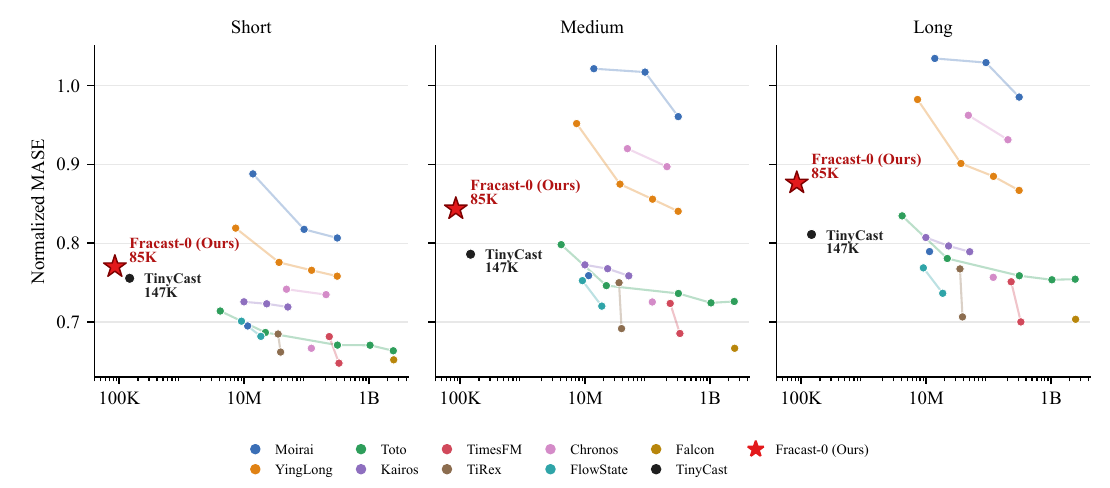}
  \caption{Normalized MASE against parameter count for the short, medium, and long horizon groups of GIFT-Eval.}
  \label{fig:shot}
\end{figure}

\subsection{Forecasting accuracy}

\paragraph{Performance and scaling.} Table~\ref{tab:main} gives the grouped scores and Figure~\ref{fig:scaling} places them in the full released-checkpoint distribution. \method{} obtains $0.808$ overall MASE and $0.564$ WQL with $85$K parameters, whereas TinyCast obtains $0.774$ and $0.545$ with $147$K parameters. Thus the $1.72\times$ larger TinyCast is $4.2\%$ better on MASE and $3.3\%$ better on WQL. Table~\ref{tab:main} also locates the WQL gap: the short-horizon values are nearly tied at $0.569$ and $0.570$, while the medium- and long-horizon values are $0.557$ and $0.556$ for \method{} versus $0.515$ and $0.515$ for TinyCast. Figure~\ref{fig:scaling}(a) sharpens the scale contrast: FlowState-r1.1, TiRex-2, and TimesFM-3 reach $0.701$, $0.678$, and $0.667$ at $18.5$M, $38.4$M, and $331$M parameters, so these improvements over \method{} cost $218\times$, $452\times$, and $3{,}891\times$ as many parameters. Figure~\ref{fig:scaling}(b) shows the same ordering on WQL ($0.487$, $0.467$, and $0.456$ versus $0.564$), but the larger systems' gains over \method{} are $13.7\%$, $17.2\%$, and $19.2\%$, far below their parameter multipliers. Table~\ref{tab:main} and these two scaling panels therefore support the bounded conclusion that \method{} is a \textbf{usable point near the extreme low-parameter end rather than the best absolute predictor}. The next development target is to close the medium- and long-horizon gap without surrendering the compact parameter budget.

\paragraph{Pareto optimality.} Figure~\ref{fig:pareto} isolates the empirical parameter--accuracy trade-off. In the overall panel, \method{} is the first of the eight MASE-frontier steps at $85$K parameters ($0.808$ MASE and $0.564$ WQL). TinyCast is the next step: its extra $62$K parameters buy $4.2\%$ lower MASE and $3.3\%$ lower WQL. The next MASE step, Toto-4M, requires another $28.3\times$ parameters for only $2.2\%$ additional MASE improvement, and the final TimesFM-3 to Falcon-2.0 edge spends another $7.5\times$ parameters for $0.1\%$. In the horizon panel, the \method{}-to-TinyCast MASE changes are $2.0\%$, $7.4\%$, and $7.5\%$ for short, medium, and long terms, showing that the smallest checkpoint is closest to its successor where prediction extends over fewer future blocks. The figure consequently establishes \textbf{non-domination and diminishing marginal returns} under this checkpoint set, but not a causal scaling law: the frontier mixes architectures, training corpora, and released schedules, and weight sharing itself remains untested without an unshared control at the same budget.

\paragraph{Horizon sensitivity.} Figure~\ref{fig:shot} explains where the aggregate gap originates. In Figure~\ref{fig:shot}(a), the short-horizon MASE values are $0.770$ for \method{} and $0.755$ for TinyCast, only a $2.0\%$ separation. Figure~\ref{fig:shot}(b) moves to medium horizons, where the values become $0.844$ and $0.786$, widening the separation to $7.4\%$. Figure~\ref{fig:shot}(c) gives long horizons at $0.877$ and $0.811$, a similarly clear $7.5\%$ gap. From short to long horizons, \method{}'s MASE increases by $13.9\%$, whereas TinyCast's increases by only $7.4\%$. This pattern is consistent with the cost of extending a $48$-step block by repeated rollout: short targets need less error accumulation, whereas medium and long targets expose it. Because each horizon panel aggregates $55$, $21$, and $21$ configurations, the figure does not isolate rollout length as the only cause. The resulting conclusion is \textbf{near short-horizon parity but weaker horizon stability beyond the short group}; the next diagnostic should evaluate medium- and long-horizon rollout before adding capacity.

\subsection{Deployment efficiency}

\begin{figure}[h]
  \centering
  \begin{subfigure}[t]{0.49\textwidth}
    \includegraphics[width=\linewidth]{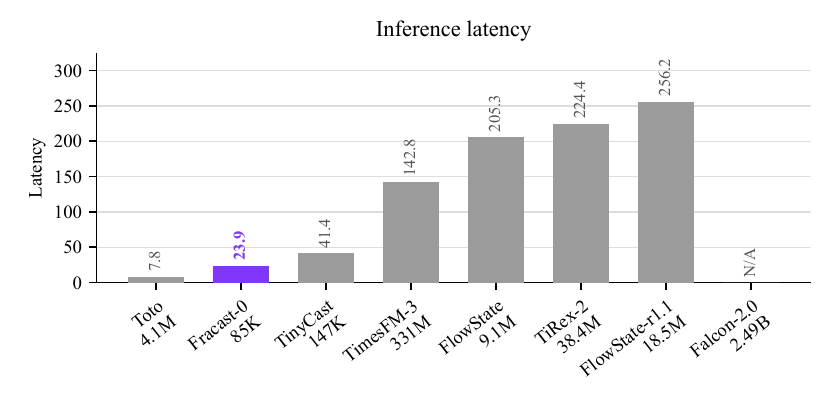}
    \caption{Batch-1 median latency (ms).}
    \label{fig:cpu-latency}
  \end{subfigure}
  \hfill
  \begin{subfigure}[t]{0.49\textwidth}
    \includegraphics[width=\linewidth]{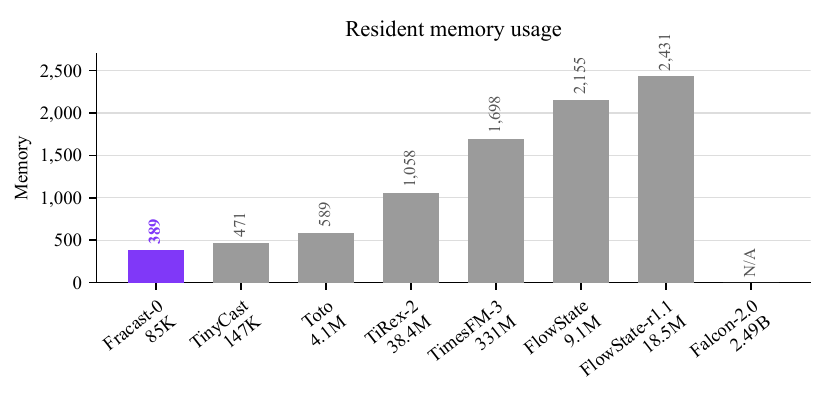}
    \caption{Resident memory (MiB).}
    \label{fig:cpu-memory}
  \end{subfigure}
  \caption{Deployment efficiency on the local CPU benchmark. Both panels use
    the same eight-model set and fixed random-input workload. The memory panel
    reports total resident memory after inference. Falcon-2.0 is N/A because no
    local checkpoint is available.}
  \label{fig:cpu-efficiency}
\end{figure}

\paragraph{Performance test.} CPU latency and resident memory were measured on a MacBook Air with an Apple M5 chip, 24 GB of unified memory, and 10 CPU cores, using PyTorch 2.14.0. All eight local checkpoints ran on CPU at batch size one under the same fixed random-input workload. Figure~\ref{fig:cpu-efficiency} tests whether the parameter reduction survives a fixed local batch-1 CPU workload. Figure~\ref{fig:cpu-latency} reports $23.9$ ms for \method{} and $41.4$ ms for TinyCast, a $42.3\%$ reduction that closely tracks the $42.0\%$ parameter reduction. Figure~\ref{fig:cpu-memory} reports $389$ versus $471$ MiB but only a $17.4\%$ reduction. This divergence indicates that runtime footprint is not parameter count alone because resident memory still depends on tensors, buffers, and framework overhead. The latency panel also bounds the claim: Toto-4M is fastest at $7.8$ ms but uses $589$ MiB, whereas TimesFM-3, TiRex-2, and FlowState-r1.1 are slower at $142.8$, $224.4$, and $256.2$ ms and require $1{,}698$, $1{,}058$, and $2{,}431$ MiB. \method{} is therefore the \textbf{lightest resident-memory point and improves the compact TinyCast baseline on both CPU metrics}, but it is not a universal fastest model. Deployment claims should remain within this local batch-1 workload until percentile latency and peak-memory distributions are reported.

\paragraph{Deployment implications.} The short-horizon parity and CPU profile favor \method{} when many local series must be scored on memory-constrained devices, especially at daily and hourly frequencies. Sub-hourly forecasting and extended rollouts remain where added capacity is more valuable, so deployment should weigh horizon, frequency, and local resource budget rather than parameter count alone. Appendix~\ref{app:qualitative} gives trajectory-level comparisons, while Appendices~\ref{app:official-domain}--\ref{app:official-variate} report domain, frequency, and variable-type breakdowns.

\flushbottom


\section{Conclusion}

We presented \method{}, an $85$K-parameter time series foundation model that
exploits temporal self-similarity by reusing one block along each dilation
ladder. Across $97$ GIFT-Eval configurations, \method{} remains non-dominated at
MASE $0.808$ and WQL $0.564$, uses $42.0\%$ fewer parameters than TinyCast, and
has $42.3\%$ lower batch-1 CPU latency. These results support cross-scale weight reuse as a viable
path to further TSFM compression rather than establish it as a general law. Our evidence is limited to one \method{} architecture and one benchmark, and
we did not run large-scale ablations because compute was limited. Future work
should test different stocks, horizons, and regimes at scale and determine which
same-level weights to merge and which to keep separate, including whether depth
still helps once weights are shared.

\FloatBarrier
\bibliographystyle{assets/plainnat}
\bibliography{refs}

\clearpage
\appendix
\section{Experiment Details}
\label{app:detail}

\subsection{Training and Evaluation Protocol}
\label{app:training-evaluation-protocol}

\paragraph{Training configuration.} \method{} holds 85{,}001 parameters with width $D=64$, ten dilations from $1$ to $512$, a context of $2{,}048$ steps, a horizon of $48$ steps, and nine quantiles. Pretraining uses AdamW with $\beta_2=0.95$, weight decay $0.01$, and gradient clipping $1.0$, with a peak learning rate of $3\times10^{-3}$, a 1{,}831-step warmup, a 65\% stable fraction, and decay to $10^{-5}$. The base schedule runs 36{,}621 steps at an effective batch of 4{,}096 windows (micro-batch 512 with 8 accumulation steps) under bf16 autocast, and the released checkpoint extends it to 40{,}075 steps, about 164M windows. The committing term carries $\lambda=0.3$, training unrolls four future chunks with scheduled sampling rising to $0.5$, and augmentation applies time flip, sign flip, down-sampling, and mixup.

\paragraph{Evaluation protocol.} GIFT-Eval defines one evaluation unit by dataset, sampling frequency, and forecast term. This gives $97$ released test configurations with $55$ short, $21$ medium, and $21$ long members. We evaluate these windows without per-dataset fine-tuning and emit nine quantile levels from $0.1$ to $0.9$. For each unit, MASE is computed from the median forecast and WQL from the nine quantiles; both are divided by the corresponding Seasonal Naive score, so lower values are better. Overall and group scores are geometric means of these normalized ratios, which gives every configuration equal weight in the aggregate. \method{} is evaluated with the local harness, while external per-configuration scores are recomputed from the GIFT-Eval Space CSVs\footnote{Revision \texttt{81100d6fc0361dc011fb984d4e22d9b71af3c3c4}, accessed 2026-09-22.}.

\subsection{Qualitative short-horizon forecasts}
\label{app:qualitative}

\begin{figure}[h]
  \centering
  \includegraphics[width=0.8\textwidth]{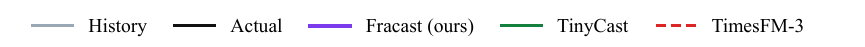}\\[-0.6ex]
  \begin{minipage}[t]{0.32\textwidth}
    \centering
    \includegraphics[width=\linewidth]{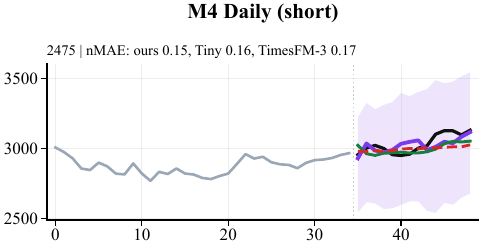}
  \end{minipage}\hfill
  \begin{minipage}[t]{0.32\textwidth}
    \centering
    \includegraphics[width=\linewidth]{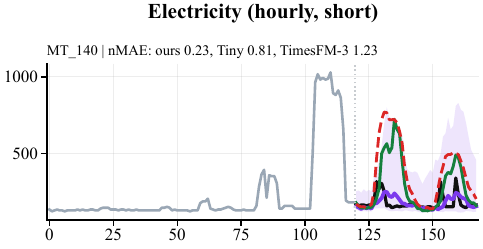}
  \end{minipage}\hfill
  \begin{minipage}[t]{0.32\textwidth}
    \centering
    \includegraphics[width=\linewidth]{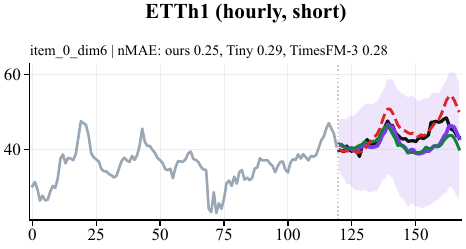}
  \end{minipage}\\[0.4ex]
  \begin{minipage}[t]{0.32\textwidth}
    \centering
    \includegraphics[width=\linewidth]{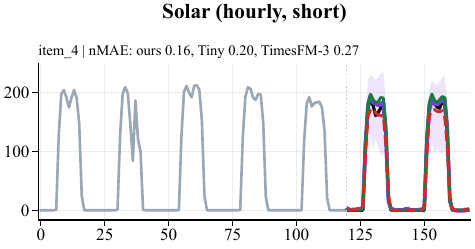}
  \end{minipage}\hfill
  \begin{minipage}[t]{0.32\textwidth}
    \centering
    \includegraphics[width=\linewidth]{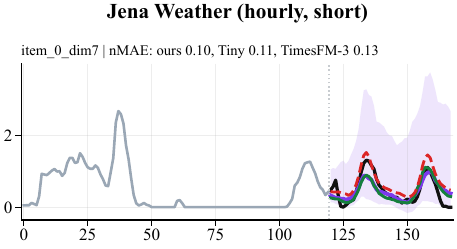}
  \end{minipage}\hfill
  \begin{minipage}[t]{0.32\textwidth}
    \centering
    \includegraphics[width=\linewidth]{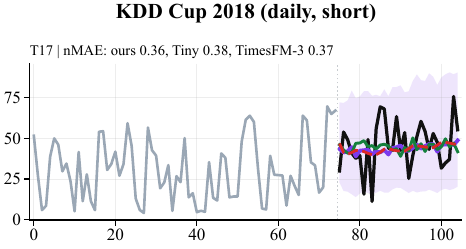}
  \end{minipage}
  \caption{Qualitative short-horizon forecasts on six GIFT-Eval datasets. The shaded band is \method{}'s 10--90\% quantile interval, and the panel title reports normalized MAE for the selected window.}
  \label{fig:qualitative}
\end{figure}

Figure~\ref{fig:qualitative} compares the three models on six selected GIFT-Eval windows. The panels span smooth daily levels, sharp load changes, daily weather cycles, periodic solar generation, and noisy event counts. \method{} has the lowest panel-title nMAE in all six windows, but the margins differ sharply. Electricity is the clearest case, where the relative gains over TinyCast and TimesFM-3 are $71.6\%$ and $81.3\%$; KDD Cup is the closest case, where the corresponding gains are $5.3\%$ and $2.7\%$. M4 Daily and KDD Cup have nearly overlapping medians, so their small score differences should not be read as visually decisive separations. In Electricity, the baselines introduce a large early peak, whereas \method{} remains closer to the observed low level before tracking the later activity. ETTh1, Solar, and Jena Weather preserve the dominant periodic timing, and the remaining score differences mainly reflect peak height. Taken together, these panels indicate that \method{}'s \textbf{clearest short-horizon advantage appears under abrupt load changes}, while its medians track the same dominant levels or cycles as the baselines elsewhere. The six cases therefore support a bounded conclusion that short-horizon tracking is competitive across these selected daily and hourly signals, but they do not estimate an aggregate win rate.

\subsection{Official By Domain comparison}
\label{app:official-domain}

Table~\ref{tab:appendix-domain-mase} reports the seven-domain view of the ten checkpoints in Table~\ref{tab:main} at GIFT-Eval Space. MASE is the Seasonal Naive-normalized geometric mean. Across domains, \method{} remains below Seasonal Naive but \textbf{does not lead any domain}. TinyCast is better on six domains, with Sales as the exception, while TiRex-2, Chronos-2, TimesFM-3, Toto-2.5B, and Falcon-2.0 provide the strongest domain-level references.

\begin{table}[h]
  \centering
  \caption{Official GIFT-Eval By Domain comparison. Each cell reports MASE.}
  \label{tab:appendix-domain-mase}
  \resizebox{\textwidth}{!}{%
    \renewcommand{\arraystretch}{1.08}%
    \setlength{\tabcolsep}{15pt}%
    \begin{tabular}{@{}l*{7}{c}@{}}
      \toprule
      Model                     & Econ/Fin & Energy & Healthcare & Nature & Sales & Transport & Web/CloudOps \\
      \midrule
      \textbf{\method{} (Ours)} & 0.909    & 0.904  & 0.756      & 0.765  & 0.692 & 0.686     & 0.805        \\
      TinyCast                  & 0.895    & 0.882  & 0.734      & 0.758  & 0.734 & 0.670     & 0.696        \\
      Moirai-2.0                & 0.779    & 0.837  & 0.600      & 0.755  & 0.689 & 0.620     & 0.665        \\
      FlowState-r1.1            & 0.756    & 0.816  & 0.597      & 0.705  & 0.682 & 0.604     & 0.628        \\
      TiRex-2                   & 0.776    & 0.767  & 0.574      & 0.705  & 0.682 & 0.576     & 0.609        \\
      Chronos-2                 & 0.772    & 0.816  & 0.551      & 0.723  & 0.679 & 0.603     & 0.611        \\
      YingLong-310M             & 0.899    & 0.870  & 0.703      & 0.746  & 0.742 & 0.666     & 0.846        \\
      TimesFM-3                 & 0.739    & 0.798  & 0.535      & 0.687  & 0.682 & 0.543     & 0.582        \\
      Toto-2.5B                 & 0.739    & 0.817  & 0.534      & 0.724  & 0.699 & 0.594     & 0.616        \\
      Falcon-2.0                & 0.694    & 0.808  & 0.563      & 0.640  & 0.688 & 0.575     & 0.575        \\
      \bottomrule
    \end{tabular}%
  }
\end{table}

\subsection{Official By Frequency comparison}
\label{app:official-frequency}

Table~\ref{tab:appendix-frequency-mase} reports the eight-frequency view of the ten checkpoints in Table~\ref{tab:main}. \method{} remains below Seasonal Naive in seven groups; only Secondly is above $1.0$, at $1.141$. It tracks TinyCast within $1.4\%$ on Daily, Hourly, and Weekly, is worse at Secondly and Minutely by $25.8\%$ and $9.2\%$, and is better on Monthly, Quarterly, and Yearly by $3.4\%$, $6.8\%$, and $12.4\%$. The comparison therefore identifies \textbf{sub-hourly forecasting as the main frequency-specific weakness}, while the longer-period groups improve over TinyCast but remain behind TimesFM-3 and Falcon-2.0.

\begin{table}[h]
  \centering
  \caption{Official GIFT-Eval By Frequency comparison. Each cell reports MASE.}
  \label{tab:appendix-frequency-mase}
  \resizebox{\textwidth}{!}{%
    \renewcommand{\arraystretch}{1.08}%
    \setlength{\tabcolsep}{15pt}%
    \begin{tabular}{@{}l*{8}{c}@{}}
      \toprule
      Model                     & Secondly & Minutely & Hourly & Daily & Weekly & Monthly & Quarterly & Yearly \\
      \midrule
      \textbf{\method{} (Ours)} & 1.141    & 0.826    & 0.765  & 0.711 & 0.869  & 0.855   & 0.801     & 0.818  \\
      TinyCast                  & 0.907    & 0.756    & 0.754  & 0.711 & 0.859  & 0.885   & 0.859     & 0.934  \\
      Moirai-2.0                & 0.854    & 0.727    & 0.710  & 0.670 & 0.769  & 0.806   & 0.739     & 0.837  \\
      FlowState-r1.1            & 0.785    & 0.697    & 0.689  & 0.672 & 0.716  & 0.772   & 0.713     & 0.754  \\
      TiRex-2                   & 0.721    & 0.684    & 0.631  & 0.654 & 0.770  & 0.784   & 0.737     & 0.892  \\
      Chronos-2                 & 0.698    & 0.700    & 0.698  & 0.653 & 0.725  & 0.756   & 0.720     & 0.816  \\
      YingLong-310M             & 1.252    & 0.786    & 0.752  & 0.711 & 0.851  & 0.867   & 0.868     & 1.087  \\
      TimesFM-3                 & 0.679    & 0.672    & 0.643  & 0.640 & 0.720  & 0.744   & 0.722     & 0.793  \\
      Toto-2.5B                 & 0.760    & 0.686    & 0.682  & 0.655 & 0.751  & 0.775   & 0.719     & 0.867  \\
      Falcon-2.0                & 0.685    & 0.663    & 0.658  & 0.619 & 0.690  & 0.779   & 0.769     & 0.832  \\
      \bottomrule
    \end{tabular}%
  }
\end{table}

\subsection{Official By Variate Type comparison}
\label{app:official-variate}

Table~\ref{tab:appendix-variate-mase} reports the multivariate and univariate view of the same ten checkpoints. \method{} obtains MASE $0.831$ on multivariate inputs and $0.790$ on univariate inputs, both below Seasonal Naive. The $0.041$ absolute gap corresponds to a $5.2\%$ relative increase, and every checkpoint in the table has higher multivariate MASE. \method{} trails TinyCast in both groups and leads neither group, so its \textbf{remaining accuracy gap is broad rather than specific to one input type}, with multivariate inputs modestly harder.

\begin{table}[h]
  \centering
  \caption{Official GIFT-Eval By Variate Type comparison. Each cell reports MASE.}
  \label{tab:appendix-variate-mase}
  \resizebox{\textwidth}{!}{%
    \renewcommand{\arraystretch}{1.08}%
    \setlength{\tabcolsep}{95pt}%
    \begin{tabular}{@{}l*{2}{c}@{}}
      \toprule
      Model                     & Multivariate & Univariate \\
      \midrule
      \textbf{\method{} (Ours)} & 0.831        & 0.790      \\
      TinyCast                  & 0.780        & 0.769      \\
      Moirai-2.0                & 0.752        & 0.709      \\
      FlowState-r1.1            & 0.717        & 0.689      \\
      TiRex-2                   & 0.704        & 0.657      \\
      Chronos-2                 & 0.709        & 0.689      \\
      YingLong-310M             & 0.840        & 0.766      \\
      TimesFM-3                 & 0.679        & 0.658      \\
      Toto-2.5B                 & 0.712        & 0.683      \\
      Falcon-2.0                & 0.683        & 0.653      \\
      \bottomrule
    \end{tabular}%
  }
\end{table}

\section{AI use statement}
\label{app:ai-use-statement}

In this work, we used generative AI tools, including large language models accessed through research, coding, and writing assistants, for literature discovery and summarization, brainstorming and refinement of research questions, conceptual framing, hypotheses, and claims, design and feedback on research methodology and experiments, implementation and debugging of model, training, data, and evaluation code, data preprocessing and translation or language editing, interpretation and presentation of experimental results, drafting and restructuring of manuscript sections, editing and polishing of manuscript prose, and preparation or modification of tables, figures, references, and other research artifacts.

We did not use generative AI tools to generate the training or evaluation datasets or to prove mathematical claims. We reviewed all AI-assisted work before using it in this paper. We checked literature claims and citations against primary papers and official repositories, verified the manuscript equations and technical descriptions manually, tested and inspected the code, and independently reproduced the reported numerical results from archived outputs. The authors made all final decisions about the research questions, method, experiments, claims, and presentation, reviewed every AI-assisted contribution.

\end{document}